\documentclass[10pt, a4paper]{article}

\usepackage{lrec2026}
\usepackage{graphicx}
\usepackage{booktabs}
\usepackage{amsmath}
\usepackage{multirow}
\usepackage{tikz-dependency}

\title{The Grammar Does the Work:\\Functional vs. Lexical Dependency Length Minimization Across Universal Dependencies}

\name{Kim Gerdes\sthanks{\ This paper was entirely conceived, written, and coded by Claude Opus 4.6 (Anthropic) in agentic mode. The author provided prompts and editorial oversight but did not originate the research idea, write code, or draft text. See the Ethics Statement and AI Disclosure section for full disclosure. All code and data are available at \url{https://github.com/typometrics/UDW26-Dependency-Length-Minimization} under a CC~BY~4.0 license.}}

\address{Universit\'e Paris-Saclay, LISN (CNRS) \\
         Orsay, France \\
         gerdes@lisn.fr}

\abstract{
Dependency length minimization (DLM) is a well-documented processing universal, but previous studies report a single mean dependency distance (MDD) per language, obscuring variation across syntactic relation types. We analyze \textbf{122 languages} in \textbf{UD} and \textbf{SUD} (version 2.17), showing that DLM operates on two distinct levels. \textbf{Grammar-driven optimization} targets functional dependencies (det, case, aux), which are universally short (mean 1.71, $\sigma$\,=\,0.33) and invariant across typologically diverse languages. \textbf{Processing-driven optimization} operates on lexical dependencies (nsubj, obj, obl), which are longer (mean 2.87), highly variable ($\sigma$\,=\,0.63), and constrained by word-order typology. This asymmetry holds in SUD despite reversed head direction ($r$\,=\,0.92). We conclude that ``the grammar does the work'' of minimization by scaffolding sentences with local functional attachments, leaving processing pressures to determine the ordering of lexical heads.
\\ \newline \Keywords{dependency length minimization, Universal Dependencies, Surface-Syntactic UD, functional dependencies, lexical dependencies, syntactic typology}}

\begin{document}

\maketitleabstract

\section{Introduction}

The tendency to minimize the linear distance between syntactically related words --- dependency length minimization (DLM) --- is one of the best-supported universals in quantitative linguistics \citep{futrell2015,temperley2018}. Within dependency grammar, \citet{hudson1995} was the first to link dependency distance with processing difficulty. \citet{gibson1998} formalized this insight, proposing that sentence processing difficulty increases with the distance between a word and the head to which it must be integrated; minimizing dependency length thus reduces working memory demands during incremental parsing. \citet{liu2008} provided the first large-scale quantitative test of the dependency distance minimization hypothesis across languages and proposed mean dependency distance (MDD) as a metric of language comprehension difficulty. This cognitive motivation has been supported by extensive cross-linguistic evidence showing that observed dependency lengths are significantly shorter than random baselines \citep{gildea2010,futrell2020}.

Despite the robustness of the aggregate DLM signal, a fundamental question remains: \emph{does DLM operate uniformly across all types of syntactic dependencies?} Previous large-scale studies report a single MDD per language, aggregating dependencies as diverse as determiners (which must be adjacent to their noun) and subjects (which can be arbitrarily far from their verb). As \citet{xliu2022} noted in a diachronic study, ``dependency distance minimization is not universal across all dependency types,'' with only a subset of relation types responsible for the observed minimization effect.

We propose that DLM is not a uniform pressure, but operates on two distinct levels, corresponding to the fundamental distinction between \emph{functional} and \emph{lexical} projections in syntactic theory \citep{tesniere1959,melcuk1988}.
\begin{enumerate}
    \item \textbf{Grammar-driven minimization:} Functional heads (determiners, case markers, auxiliaries) are closed-class items whose position is strictly constrained by grammatical linearization rules. These rules ``hard-code'' minimization by mandating adjacency (e.g., Det adjacent to Noun).
    \item \textbf{Processing-driven minimization:} Lexical dependencies (subjects, objects, modifiers) involve open-class elements whose ordering is more flexible. Here, minimization is a soft constraint competing with information structure and other communicative needs.
\end{enumerate}

We test this hypothesis on \textbf{122 languages} (all UD/SUD v2.17 languages with $\geq$500 sentences; see \S3.1) in both \textbf{UD} and \textbf{SUD}. We concatenate all treebanks per language to create a representative sample. This dual-framework comparison is methodologically important: \citet{osborne2019} showed that UD's content-word-headed convention inflates MDD, as function words are treated as dependents of distant lexical heads rather than as local heads themselves; converting to syntactic structures where function words head their phrases significantly reduces MDD. By contrasting UD and SUD, we disentangle annotation effects from processing patterns.

\section{Related Work}

\subsection{Dependency Length Minimization}

DLM has a rich empirical history. \citet{liu2008} proposed MDD as a metric of language comprehension difficulty and was the first to test the DLM hypothesis quantitatively across languages; we note that MDD (the mean of per-dependency distances) differs from the dependency length (DL) sum used by \citet{futrell2015} \citep[see][for a detailed discussion]{niu2025}. \citet{temperley2008} identified three principles that minimize dependency length: consistent branching direction, shorter dependent phrases being closer to the head, and opposite-branching of one-word phrases. \citet{gildea2010} confirmed that English dependency lengths are much closer to optimal than to random.

\citet{futrell2015} scaled this to 37 languages, demonstrating universal DLM, and \citet{temperley2018} framed DLM as a ``typological/cognitive universal''. \citet{ferrer2022} developed an optimality score framing word order as a spatial network optimization problem. More recently, \citet{futrell2020} showed that dependency locality accurately predicts word-order preferences.

\subsection{Dependency Types and DLM}

Most critically for our work, a few studies have considered whether DLM varies across dependency types. \citet{xliu2022} examined diachronic changes in dependency distance by relation type in English, finding that only 9 types are responsible for overall minimization (including \texttt{aux}, \texttt{mark}, \texttt{nsubj}, and \texttt{ccomp}), while 6 types actually \emph{increased} in distance over time (including \texttt{det}, \texttt{amod}, and \texttt{compound}). Crucially, their study measures \emph{diachronic trend direction} --- whether distances got shorter or longer across centuries --- not absolute shortness. Their 9 minimizing types mix functional (\texttt{aux}, \texttt{mark}) and lexical (\texttt{nsubj}, \texttt{ccomp}) relations, because diachronic trends in English need not align with the synchronic functional/lexical distinction that holds \emph{across languages}.

\citet{dyer2023} used a parallel corpus of 35 languages to revisit DLM, finding a ``markedly lesser extent'' of minimization in verb-final languages --- an asymmetry we replicate and attribute to the lexical dependency component (\S4.3): verb-final languages display higher lexical MDD while functional MDD remains uniformly low. \citet{gao2024} used per-relation dependency distances to study syntactic complexity in Alzheimer's disease, finding that specific relation types like adverbial modifiers show differential patterns. \citet{krielke2024} showed that both scientific English and German increasingly utilize short, intra-phrasal dependency relations while long dependencies (clausal embeddings) become less favored over time --- hinting at a functional/lexical split, though not explicitly framed as such.

However, \textbf{no previous study has systematically classified dependencies into functional and lexical categories and compared their DLM behavior at scale}. Our contribution is the theoretically motivated, \emph{a priori} classification of relations into functional and lexical types, applied \emph{synchronically} across 122 languages, showing that the absolute distance gap between these categories is universal, not a language-specific historical trend.

\subsection{UD, SUD, and the Status of Function Words}

The treatment of function words is central to our analysis. Universal Dependencies \citep{demarneffe2021} adopts a content-word-headed approach where function words (determiners, auxiliaries, adpositions) are dependents of lexical heads \citep{nivre2016}. \citet{osborne2019} critiqued this convention, showing that UD's subordination of function words produces inflated MDD values compared to more syntactically motivated structures. They reported that MDD was ``significantly reduced for nearly all languages'' when converting from UD to purely syntactic structures.

Surface-Syntactic UD \citep[SUD;][]{gerdes2018,gerdes2021} addresses this by promoting function words to head status where distributionally motivated: auxiliaries govern their verbs, adpositions govern their complements, complementizers govern their clauses. This reversal provides a natural test of robustness: since $|pos(head) - pos(dep)|$ is symmetric, the same word pair produces the same distance regardless of which element is labeled head. If the functional--lexical asymmetry is real, it must hold across both annotation conventions.

\subsection{Cognitive Models and DLM}

The cognitive basis of DLM is rooted in memory constraints. \citet{gibson1998} proposed that both storage cost (keeping incomplete dependencies in memory) and integration cost (connecting incoming words to their heads) increase with dependency distance. \citet{collins2014} showed that DLM is complementary to information density optimization, suggesting that multiple cognitive pressures simultaneously shape word order. \citet{stempniak2024} further explored how DLM interacts with specific syntactic structures (coordination) in head-final languages, finding that dependency structure choices are driven by length minimization considerations. Our two-level model aligns with this: functional attachment has negligible integration cost (always local), while lexical attachment is the primary driver of processing difficulty.

\section{Data and Methodology}

\subsection{Treebank Selection}

We analyze all treebanks from UD v2.17 \citep{ud2025}. To ensure validity, we aggregate data at the language level: for each language, we concatenate all treebanks into a single corpus. We exclude languages with fewer than 500 sentences.\footnote{A bootstrap stability analysis confirms that the functional--lexical gap is robust from as few as 100 sentences (see \S\ref{sec:sensitivity}). This threshold excludes 64 low-resource languages (e.g., Guarani, Manx, Sanskrit), leaving 122 languages.} This yields a matched set of \textbf{122 languages} in \textbf{UD} and \textbf{SUD}, encompassing over 25 language families with major representation from Indo-European, Uralic, Afro-Asiatic, Tupian, Turkic, and Sino-Tibetan.
For computational efficiency on very large languages, we cap the analysis at 15{,}000 sentences per language, which provides ample data for statistical stability.
After filtering and capping, the UD dataset comprises 798{,}381 sentences and 11.2M non-punctuation dependency tokens across 122 languages (median 3{,}444 sentences per language; range 502--15{,}000). Of these tokens, 33\% are functional dependencies and 67\% are lexical, though the proportion varies considerably across languages (4\%--49\% functional), reflecting differences in morphological synthesis and the prevalence of function words.

\subsection{Functional vs. Lexical Classification}

Following the UD distinction between function words and content words \citep{nivre2016,demarneffe2021}, we classify dependency relations into two groups:

\begin{itemize}
\item \textbf{Functional}: \texttt{det}, \texttt{case}, \texttt{aux}, \texttt{mark}, \texttt{cop}, \texttt{cc}, \texttt{clf} (and subtypes). These are closed-class dependencies that mark grammatical function.

\item \textbf{Lexical}: \texttt{nsubj}, \texttt{obj}, \texttt{iobj}, \texttt{obl}, \texttt{nmod}, \texttt{amod}, \texttt{advmod}\footnote{Adverbs show mixed functional/lexical behavior. We follow UD in classifying \texttt{advmod} as lexical. A sensitivity analysis reclassifying it as functional actually strengthens the functional--lexical distinction, confirming the result's robustness (see \S\ref{sec:sensitivity}).}, \texttt{advcl}, \texttt{acl}, \texttt{xcomp}, \texttt{ccomp}, \texttt{conj}, \texttt{compound}, \texttt{appos}, \texttt{flat}, \texttt{parataxis}, etc.
\end{itemize}

\noindent\textbf{SUD adaptation.} In SUD, function words become heads \citep{gerdes2018,osborne2019}, so the label inventory changes (see Table~\ref{tab:sud_mapping} in Appendix~\ref{sec:appendix}).  The key disambiguation concerns \texttt{comp:obj} and \texttt{comp:obl}: when the head is an adposition (UPOS = ADP) or complementizer (UPOS = SCONJ), the dependency is classified as \emph{functional}; when the head is a verb, it is classified as \emph{lexical}. Of 340k \texttt{comp:*} tokens across all SUD treebanks, 62.5\% are classified functional and 37.5\% lexical.
Relations typically analyzed as non-dependency structure (e.g., \texttt{punct}, \texttt{root}) and underspecified relations (\texttt{dep}, \texttt{orphan}) are excluded.

\subsection{Metrics}
For each group (overall, functional, lexical), we compute:
\begin{enumerate}
\item \textbf{MDD}: Mean absolute distance $|pos(head) - pos(dep)|$ over non-punctuation tokens, excluding root dependencies \citep{liu2017}.
\item \textbf{Random baseline}: To estimate the expected distance under no DLM pressure, we randomly permute the \emph{linear positions} of all non-punctuation tokens in a sentence while keeping the dependency tree structure (i.e., who depends on whom) fixed. Each permutation reassigns every token to a new position, so the same tree is linearized in a different random order; dependency distances are then recomputed on this shuffled linearization. Following \citet{futrell2015}, we generate 20 such random permutations per sentence and average the resulting MDD across all permutations ($\text{MDD}_\text{rand}$). This serves as a null hypothesis: the expected distance if word order carried no DLM signal.
\item \textbf{Optimality ratio} (OR): $\text{MDD}_\text{obs} / \text{MDD}_\text{rand}$, following the formalization of \citet{ferrer2022}. This provides a normalized measure of optimization: an OR of 1.0 suggests a language is no more optimized than chance, while values approaching 0.0 indicate extreme minimization.
\item \textbf{Head Directionality}: The proportion of dependencies where the head follows the dependent ($pos(head) > pos(dependent)$).
\end{enumerate}

\section{Results}

\begin{figure*}[t!]
\centering
\includegraphics[width=0.9\textwidth]{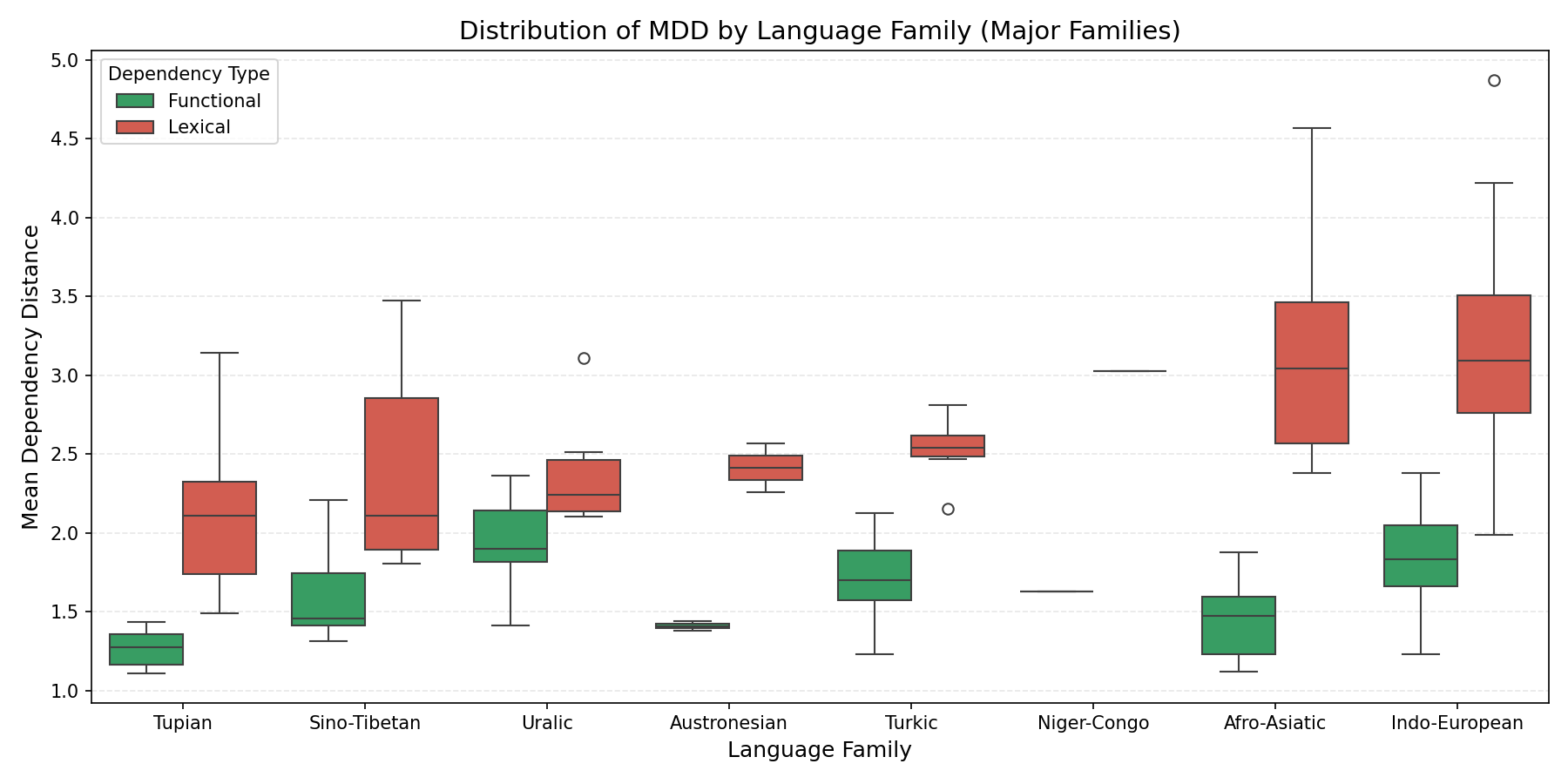}
\caption{Distribution of Functional vs.\ Lexical MDD across major language families. Functional MDD is consistently low across diverse families, whereas Lexical MDD varies significantly with word order typology (e.g., higher in head-final Turkic/Uralic/Dravidian, lower in head-initial Austronesian/Niger-Congo).}
\label{fig:family_boxplot}
\end{figure*}

\begin{figure*}[t!]
\centering
\includegraphics[width=0.9\textwidth]{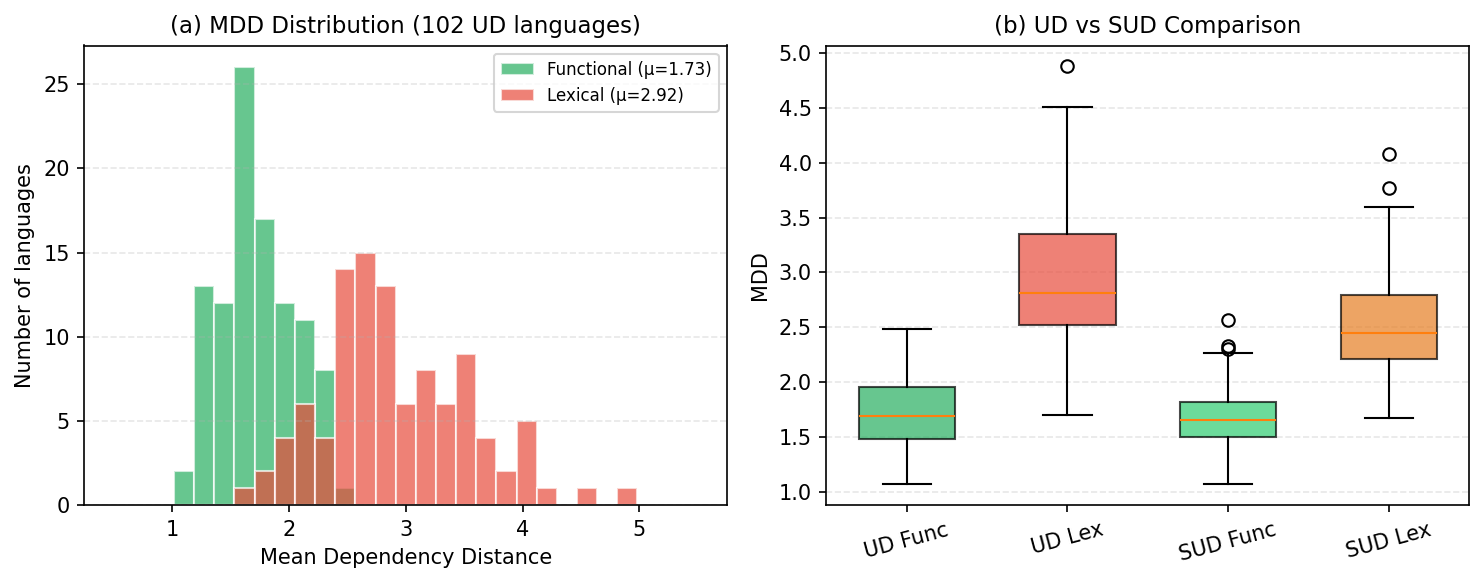}
\caption{(a) Distribution of functional (green) and lexical (red) MDD across 122 UD languages. Functional MDD clusters tightly around 1.71; lexical MDD is higher and more dispersed. (b) Boxplot comparison across UD and SUD frameworks: the functional--lexical gap is preserved regardless of annotation convention.}
\label{fig:histogram}
\end{figure*}

\subsection{Overall DLM Confirmation}

All 122 UD languages exhibit strong DLM. Observed MDD ranges from 1.44 to 3.67, far below random baselines (optimality ratios 0.17--0.89, mean 0.41). This confirms \citet{futrell2015} at scale and extends the finding to new languages.

\subsection{The Functional--Lexical Asymmetry}

\begin{figure*}[t!]
\centering
\begin{minipage}{0.48\textwidth}
\centering
\includegraphics[width=\linewidth]{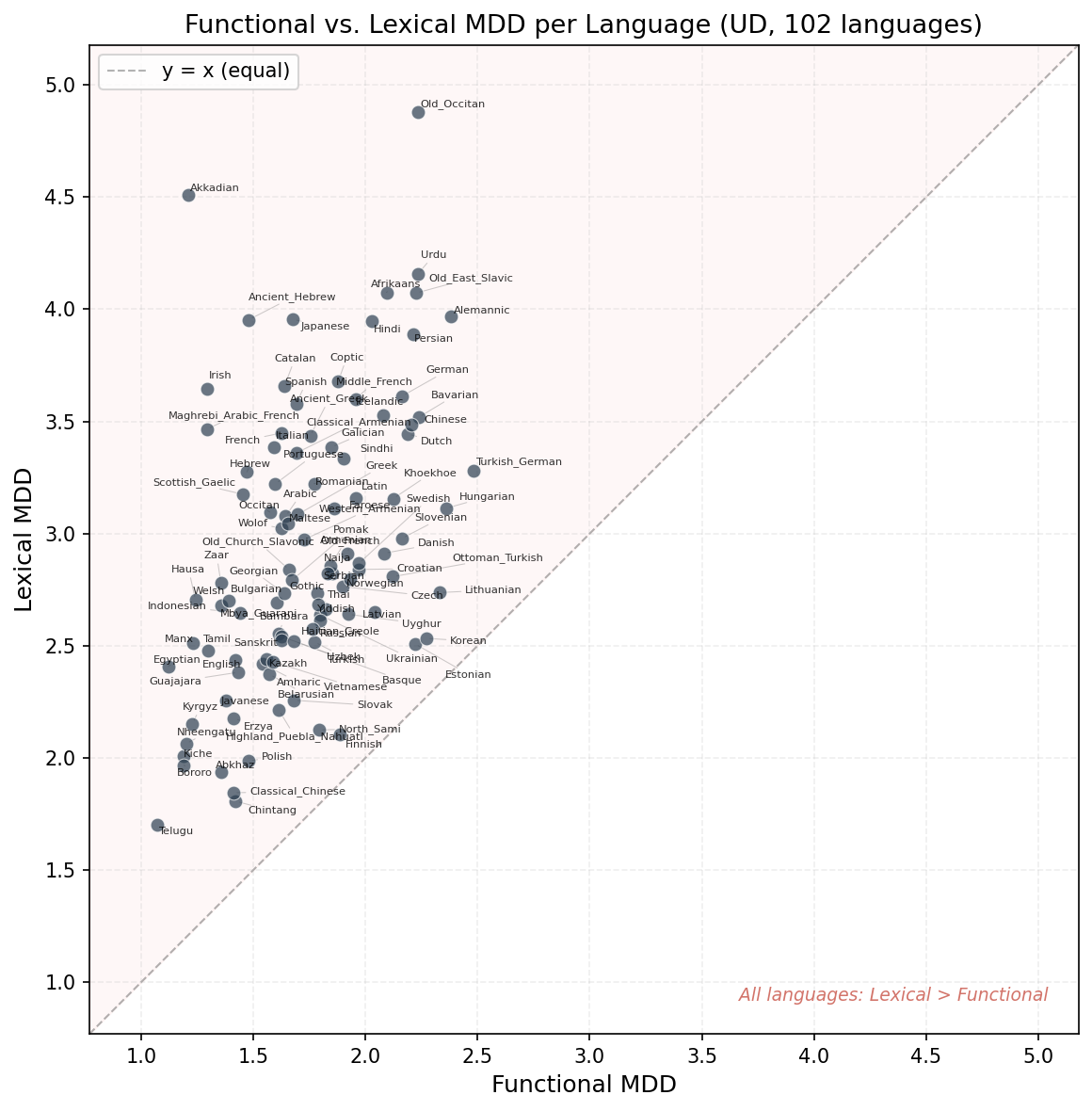}
\caption{(a) Functional vs.\ lexical MDD per language (122 UD languages). Every language is above the diagonal ($y=x$).}
\label{fig:func_vs_lex}
\end{minipage}\hfill
\begin{minipage}{0.48\textwidth}
\centering
\includegraphics[width=\linewidth]{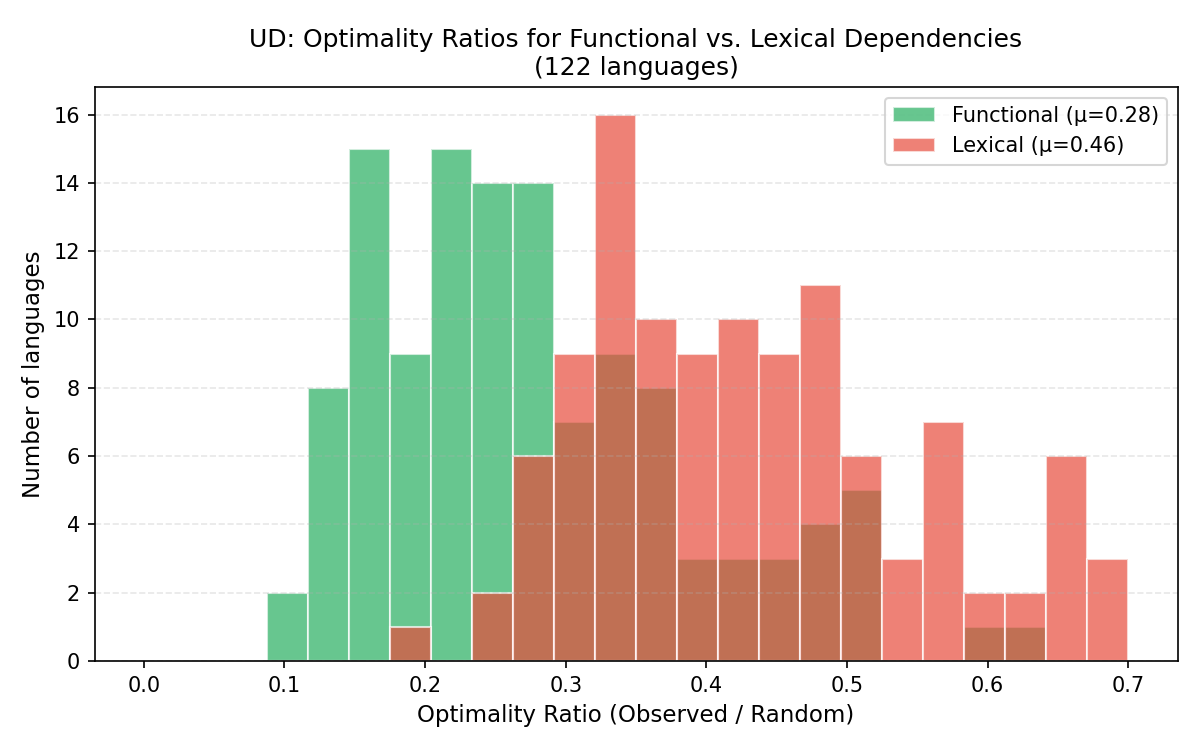}
\caption{(b) Optimality ratio distributions for functional and lexical dependencies across 122 UD languages. Lower = more optimized.}
\label{fig:opt_dist}
\end{minipage}
\end{figure*}

\begin{table}[!ht]
\centering
\scriptsize
\begin{tabular}{l|cc|cc}
 & \multicolumn{2}{c|}{\textbf{Functional}} & \multicolumn{2}{c}{\textbf{Lexical}} \\
\textbf{Framework} & MDD & OR & MDD & OR \\
\midrule
UD (122) & 1.71{\tiny$\pm$0.33} & 0.28 & 2.87{\tiny$\pm$0.63} & 0.46 \\
SUD (122) & 1.65{\tiny$\pm$0.32} & 0.27 & 2.48{\tiny$\pm$0.47} & 0.41 \\
\bottomrule
\end{tabular}
\caption{Mean ($\pm$ std) functional and lexical MDD across 122 qualifying languages per framework. Functional MDD is universally lower and less variable.}
\label{tab:funlex_summary}
\end{table}

Table~\ref{tab:funlex_summary} presents the central result, showing both absolute MDD and the optimality ratios derived from the 20-permutation random baselines. Three patterns emerge:

\textbf{1. Functional MDD is universally low.} Across 122 UD languages, functional MDD averages 1.71 with a standard deviation of only 0.33. This narrow distribution (Figure~\ref{fig:histogram}) shows that grammars universally constrain function words to appear adjacent to their hosts, regardless of language family or word-order type.

\textbf{2. Lexical dependencies also show strong DLM.} Crucially, lexical dependencies are not merely ``less optimized than functional'': with a mean optimality ratio of 0.46 ($\sigma$ = 0.15), they are 54\% shorter than random baselines in every single language (122/122, OR range 0.20--0.93). This confirms that genuine processing-driven minimization operates on lexical dependencies --- subjects, objects, and modifiers are placed substantially closer to their heads than chance would predict. However, lexical MDD is more variable ($\sigma$ = 0.63 vs.\ 0.33), and SOV and V2 languages show higher values, consistent with \citet{dyer2023}. Figure~\ref{fig:func_vs_lex} confirms that lexical MDD universally exceeds functional MDD.

\textbf{3. The two levels differ in optimization depth.} The mean functional optimality ratio is 0.28, versus 0.46 for lexical --- both well below chance, but functional OR is 39\% lower. This gap reflects different optimization mechanisms: functional adjacency is categorically enforced by grammar, while lexical ordering is a softer, gradient optimization that competes with information structure, heaviness, and other communicative pressures. The boxplot in Figure~\ref{fig:histogram}b confirms the pattern holds across both UD and SUD.

\textbf{Statistical confirmation.} We verify the functional--lexical gap with three tests, each addressing a different concern.
\emph{Is the gap consistent across languages?} A paired Wilcoxon signed-rank test compares functional and lexical MDD within each language and asks whether one is systematically lower. The result ($W = 0$, $p < 10^{-18}$) confirms that functional MDD is lower in every single language, not just on average. The effect is large: Cohen's $d = 2.30$, meaning the gap exceeds two standard deviations.
\emph{Could the gap be driven by shared genealogical history?} Related languages may share similar MDD patterns, inflating apparent universality. A linear mixed-effects model with Language Family ($N \geq 25$) as a random intercept controls for this: the functional--lexical distinction remains highly significant ($p < 0.001$), confirming the gap holds within families, not just across them.
All results replicate in SUD ($W = 0$, $p < 10^{-18}$, $d = 2.04$).\footnote{A Pitman-Morgan test further confirms that functional MDD is not merely lower but also less \emph{variable} across languages than lexical MDD ($t(120) = 8.84$, $p < 0.001$; SUD: $t = 4.96$), consistent with the claim that grammar enforces a narrow range of functional distances while lexical distances vary with typology.} Figure~\ref{fig:family_boxplot} visualizes this stability: while Lexical MDD (red) varies between head-initial and head-final families, Functional MDD (green) remains low.

\subsection{Interaction with Head Directionality}

\begin{figure*}[tbp!]
\centering
\includegraphics[width=0.9\textwidth]{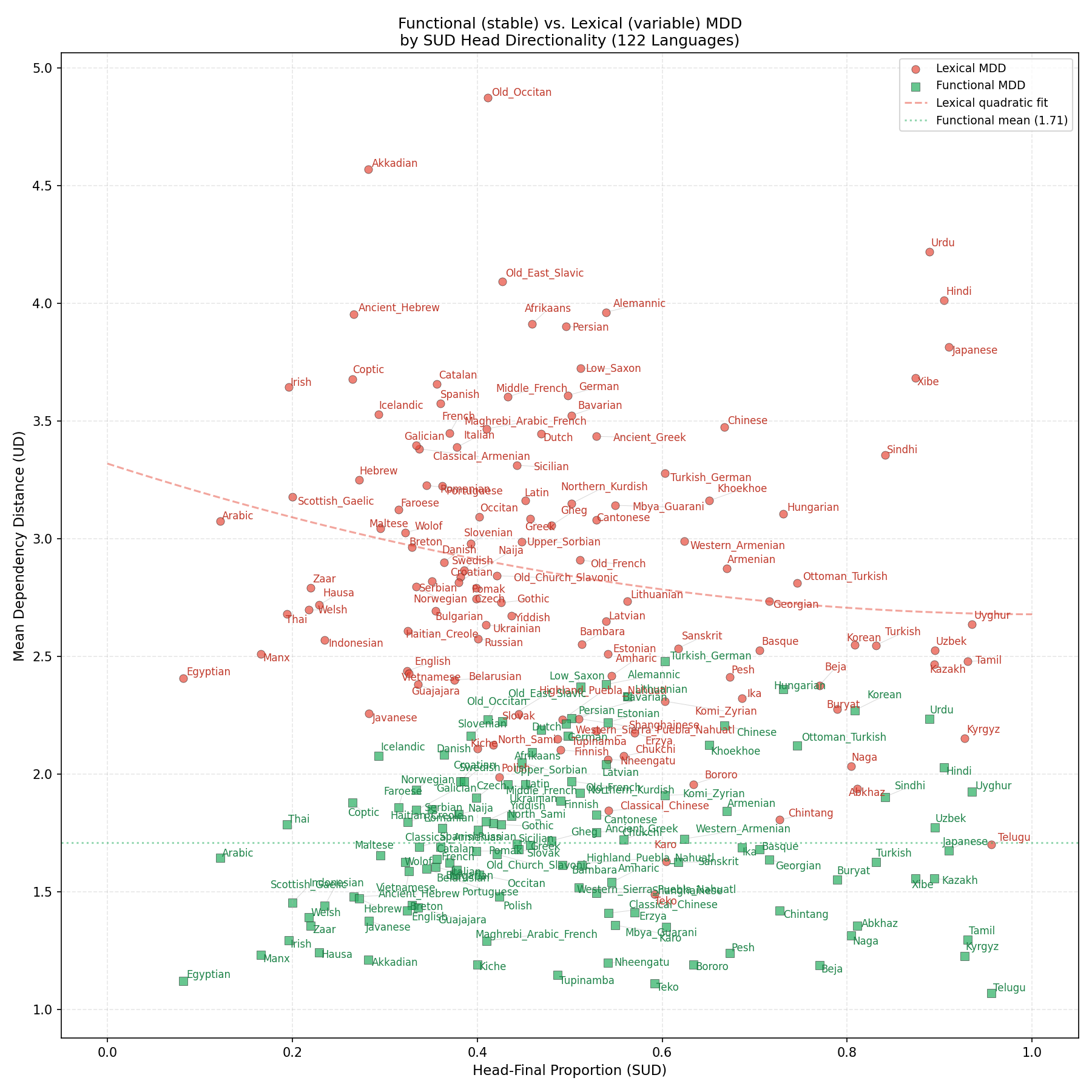}
\caption{Functional vs.\ lexical MDD in UD compared against head directionality measured in SUD. By using SUD's head-final proportion (x-axis), we capture syntactic word order (e.g., Japanese at 0.91) without the distortions caused by UD's annotation of function words.}
\label{fig:scatter}
\end{figure*}

Figure~\ref{fig:scatter} plots functional and lexical MDD against head-final proportion. Functional MDD forms a flat band around 1.71, showing negligible correlation with head directionality. Lexical MDD shows substantially more spread, with SOV languages displaying higher values. This parallels the findings of \citet{liu2010} on dependency direction as a typological parameter, but reveals that the typological signal resides exclusively in lexical dependencies.

\textbf{VO/OV classification.}
To complement the continuous head-directionality measure, we classify languages into discrete word-order types using a corpus-internal VO score, following \citet{faghiri2026}: for each VERB, we count \texttt{obj} dependencies with NOUN/PROPN dependents; the VO score is the proportion appearing to the right of the verb. Languages with VO\,$>$\,0.666 are classified as VO, those with VO\,$<$\,0.333 as OV, and the remainder as ``no dominant order'' (NDO). Of the 122 qualifying languages, 69 are VO, 35 OV, and 18 NDO.

\begin{figure*}[t!]
\centering
\includegraphics[width=0.92\textwidth]{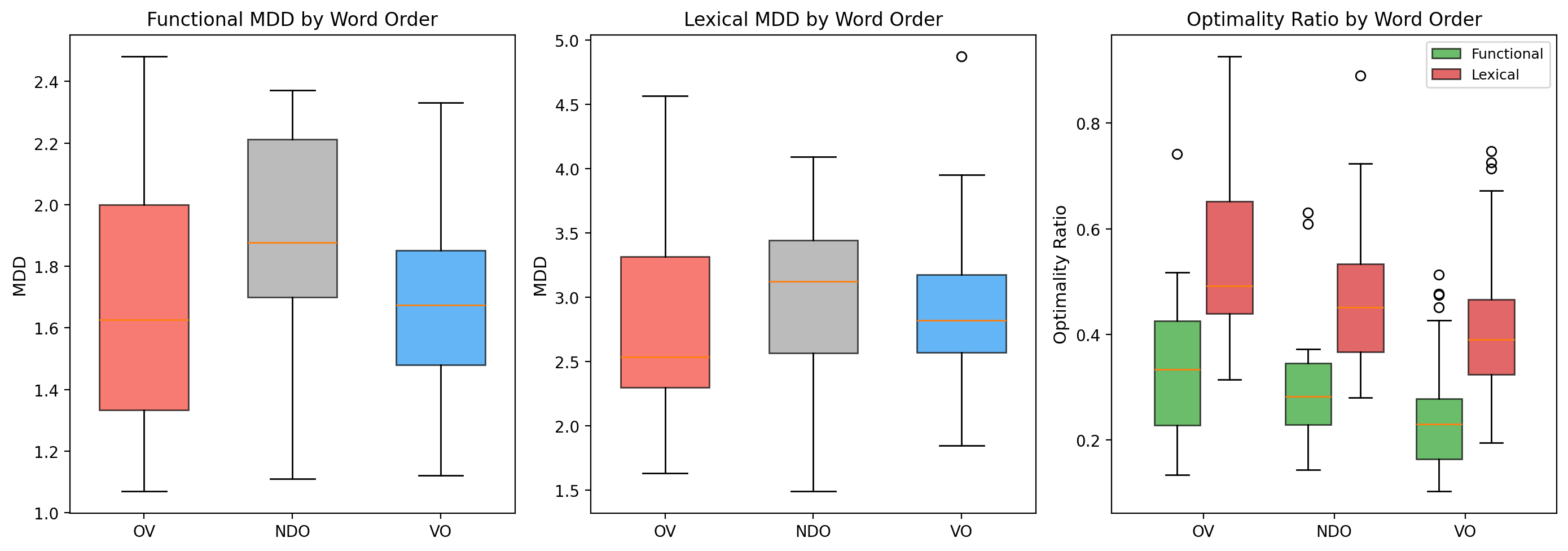}
\caption{Functional MDD (a), Lexical MDD (b), and Optimality Ratios (c, green = functional, red = lexical) by VO/OV word-order type. Absolute MDD is similar across types, but OV languages show higher optimality ratios (less room for optimization) in both categories. The functional--lexical gap is invariant ($p = 0.44$).}
\label{fig:vo_boxplots}
\end{figure*}

Figure~\ref{fig:vo_boxplots} reveals a striking pattern: absolute MDD does \emph{not} differ significantly across word-order types (functional: $p = 0.057$; lexical: $p = 0.26$), but \emph{optimality ratios} diverge sharply. OV languages have higher ORs in both functional (0.34 vs.\ 0.25) and lexical (0.55 vs.\ 0.41) categories (both $p < 10^{-4}$, Mann-Whitney). This means OV languages achieve roughly the same absolute dependency distances as VO languages, despite having less room for optimization --- their random baselines are lower because verb-final structures inherently constrain word order more tightly. Crucially, the \emph{functional--lexical gap} ($\text{Lex\_MDD} - \text{Func\_MDD}$) is invariant across types: VO = 1.19, OV = 1.12, NDO = 1.11 (Kruskal-Wallis $H = 1.65$, $p = 0.44$). This confirms that the two-level DLM asymmetry is \textbf{universal across word-order typology}.

\subsection{UD vs. SUD: Robustness Across Frameworks}

\begin{figure*}[t!]
\centering
\includegraphics[width=0.9\textwidth]{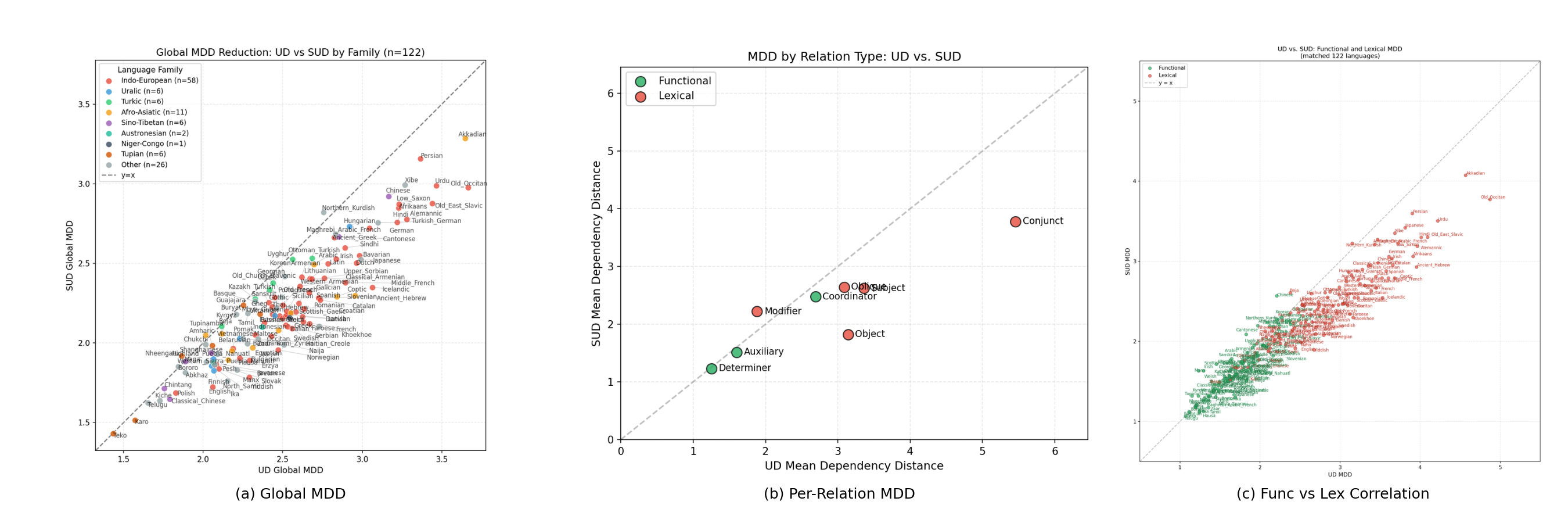}
\caption{UD vs.\ SUD comparison across 122 languages. (a) Global MDD: most languages fall below the diagonal, showing SUD lowers MDD. (b) Per-relation MDD: functional relations (green) cluster short in both frameworks. (c) Per-language functional (green) and lexical (red) MDD correlate highly across frameworks ($r > 0.92$).}
\label{fig:global_reduction}
\end{figure*}

Figure~\ref{fig:global_reduction} compares functional and lexical MDD between the frameworks. The correlations are $r = 0.92$ for functional MDD and $r = 0.92$ for lexical MDD. SUD lexical MDD is systematically lower (2.48 vs.\ 2.87). This reduction is primarily structural: in UD, oblique arguments are attached to the verb via the noun (Verb $\to$ Noun), spanning the adposition. In SUD, they are attached via the adposition (Verb $\to$ Adposition), which is typically closer to the verb than the noun is. Critically, the asymmetry between functional and lexical is preserved.

At the global level, nearly all languages fall below the diagonal (Figure~\ref{fig:global_reduction}a), confirming that SUD's head-direction choices lower global dependency distance relative to UD, as predicted by \citet{osborne2019}. Figure~\ref{fig:global_reduction}b shows per-relation MDD: functional relations (green squares) cluster in the bottom-left, indicating they are short in both frameworks.

\subsection{Per-Relation Detail}

\begin{figure*}[t!]
\centering
\includegraphics[width=0.9\textwidth]{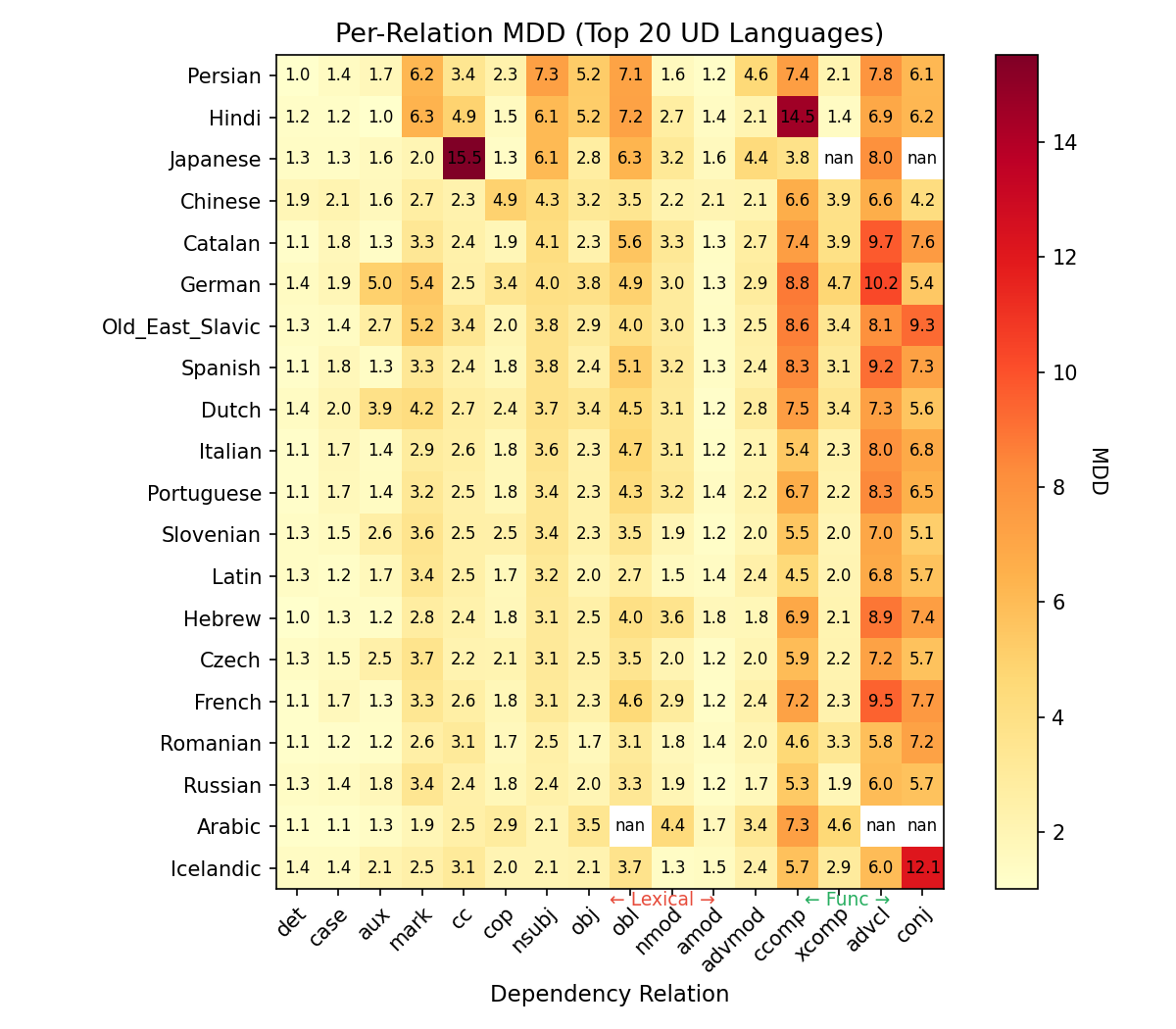}
\caption{Per-relation MDD across the 20 largest UD languages. Functional relations (det, case, aux, mark) are universally short; lexical relations (nsubj, obl, ccomp, advcl) vary with word-order typology and show higher distances.}
\label{fig:heatmap}
\end{figure*}

Figure~\ref{fig:heatmap} provides a per-relation breakdown for the 20 largest UD languages, covering 16 dependency types. Within the functional group, \texttt{det} ($\sim$1.0--1.5) and \texttt{case} ($\sim$1.0--1.8) are universally short. Within the lexical group, \texttt{nsubj} shows significant variation (1.8 in Finnish to 6+ in Hindi), reflecting SOV vs.\ SVO order. Clausal complements (\texttt{ccomp}, \texttt{advcl}) consistently show high MDD. This extends the 9-type finding of \citet{xliu2022}: the types responsible for DLM are precisely functional, while those showing variability are lexical.

\subsection{Sensitivity Analysis}\label{sec:sensitivity}

To test robustness, we recomputed the functional--lexical split under three alternative relation groupings and two alternative distance metrics (details in Appendix~\ref{sec:appendix}). Excluding \texttt{conj} (whose high MDD is an artifact of UD's chain analysis), restricting to core syntactic relations only, or applying the strictest possible classification all preserve the asymmetry: func~$<$~lex holds in at least 121/122 languages under every scenario, and the effect size remains large ($d \geq 1.65$). Additionally, excluding pronoun dependents from the lexical class \emph{widens} the gap, and computing distances in nucleus positions only (counting content words only) preserves it. A bootstrap analysis confirms the gap is detectable from as few as 100 sentences. We conclude that the finding is robust to reasonable variation in both relation classification and distance metric.

\section{Discussion}

Our results disentangle two components of the universal DLM signal reported by \citet{temperley2018} and \citet{futrell2015}. \textbf{Grammar-driven functional optimization}: functional elements are positioned adjacent to their hosts by grammatical rules \citep{temperley2008}; the universally low functional MDD (1.71 $\pm$ 0.33) is a predictable consequence of grammar design, not speaker choice. \textbf{Processing-driven lexical optimization}: this is the more revealing finding. Lexical dependencies --- where speakers have genuine ordering flexibility --- nonetheless show strong minimization in every language (OR = 0.46, i.e., 54\% shorter than random). This confirms that DLM is not merely an artifact of grammatical adjacency: even when function words are factored out, speakers consistently place arguments and modifiers closer to their heads than chance would predict. The degree of lexical minimization varies with typological constraints (OV languages show weaker optimization; \citealp{dyer2023}), suggesting that word-order typology modulates but does not eliminate processing-driven DLM. The VO/OV classification (\S4.3) makes this precise: OV languages achieve similar \emph{absolute} MDD as VO languages (2.80 vs.\ 2.88), but their higher optimality ratios (0.55 vs.\ 0.41) indicate less room for optimization given their more constrained word order. The functional--lexical gap, however, remains invariant ($p = 0.44$), confirming the two-level distinction is orthogonal to word-order typology.

This two-level view refines Gibson's Dependency Locality Theory \citep{gibson1998}: integration cost is concentrated at lexical heads, where ordering is flexible and information density effects operate \citep{collins2014}. For typology, functional MDD serves as a stable baseline ($\sigma$ = 0.33) for cross-linguistic comparison, while lexical MDD captures genuine typological variation correlating with word-order parameters \citep{liu2010,futrell2020}.

\textbf{Coordination and discourse relations.} Excluding \texttt{conj} from the lexical class does not alter the core finding (\S\ref{sec:sensitivity}). The high MDD of \texttt{conj} (5.45) is largely an artifact of UD's chain analysis; similarly, \texttt{parataxis} (6.63) and \texttt{discourse} (3.12) reflect discourse structure rather than syntactic placement. Excluding these (Scenario~C) yields a cleaner measure while preserving the functional--lexical asymmetry.

\textbf{Functional dependencies and the flux.} Our results can also be viewed through the lens of dependency flux \citep{kahane2017}. Since functional arcs are almost always short, they contribute minimally to inter-word flux: only 22.6\% of functional arcs span over any lexical arc. \texttt{det} and \texttt{case} rarely cover a lexical arc (14\%), while \texttt{mark} does so 49\% of the time --- reflecting the high \texttt{mark} MDD in SOV languages (Urdu 7.67, Hindi 6.34, German 5.38). In the flux perspective, functional dependencies contribute mostly to bouquet-local flux, while lexical dependencies drive inter-word flux complexity.

\textbf{Limitations and Future Work.} Treebank sizes and genres vary, and specific properties of individual UD treebanks (e.g., genre composition, annotation consistency, and domain) may influence observed dependency length patterns; our aggregation across treebanks per language mitigates but does not eliminate such effects. Tokenization affects absolute MDD values \citep{lei2020}. The functional/lexical boundary is theory-dependent; \texttt{advmod} is classified as lexical but could be argued either way. Future work should incorporate genre-stratified analysis, natively annotated SUD treebanks, and a full flux decomposition \citep{kahane2017} separating functional and lexical contributions to inter-word flux complexity.

\section{Conclusion}

Across \textbf{122 languages}, we find that dependency length minimization is not a monolithic phenomenon but a composite of two distinct forces.
\textbf{The grammar does the work} --- but processing does too. Functional dependencies are \emph{grammatically minimized}: by mandating local attachment for functional items (det, case, aux), the grammar guarantees a baseline of low aggregate MDD ($\approx$ 1.71), effectively scaffolding sentences with short dependencies.
\textbf{Lexical DLM is real and substantial.} When we isolate lexical dependencies --- where speakers have genuine ordering choices --- they are still 54\% shorter than random baselines across all 122 languages (OR = 0.46). This is the more informative finding: it demonstrates that online processing pressures actively shape word order beyond what grammar dictates. The variation in lexical MDD ($\approx$ 2.87, $\sigma$ = 0.63) tracks typological parameters (SOV structures incur longer distances), revealing the interplay between typological constraints and processing optimization.
This two-level view refines the efficiency-grammar hypothesis. Grammar \emph{crystallizes} minimization for the most frequent, predictable elements (function words). But the residual lexical signal shows that processing optimization operates independently and universally, even in typologically constrained languages. Future work should investigate the interaction between lexical DLM and information-structural preferences, and whether the degree of lexical optimization correlates with psycholinguistic measures of processing difficulty.

\appendix
\section{Supplementary Material}\label{sec:appendix}

\subsection{UD and SUD Dependency Example}

\begin{figure*}[!t]
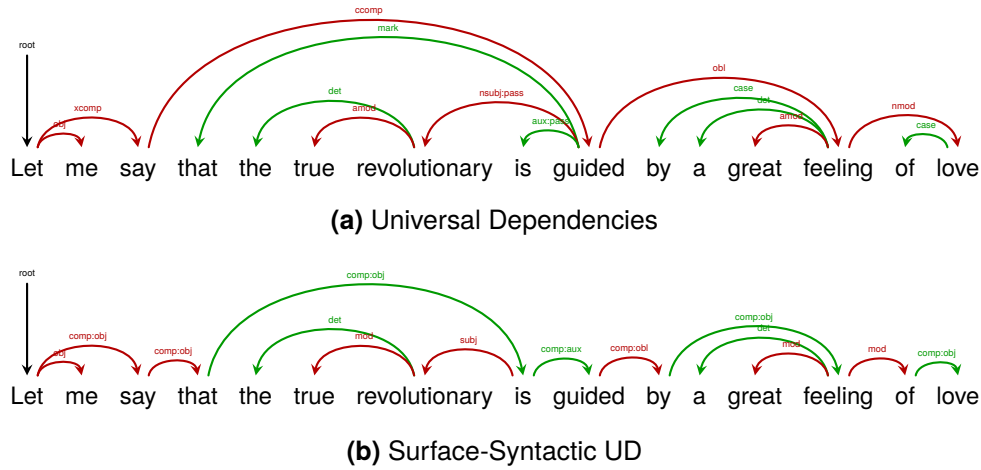

\centering
\begin{dependency}[theme=simple, edge style={thick}, label style={font=\tiny}]
\begin{deptext}[column sep=0.1cm, font=\small]
Let \& me \& say \& that \& the \& true \& revolutionary \& is \& guided \& by \& a \& great \& feeling \& of \& love \\
\end{deptext}
\deproot[edge unit distance=2ex]{1}{root}
\depedge[edge style={red!70!black, thick}, label style={text=red!70!black}]{1}{2}{obj}
\depedge[edge style={red!70!black, thick}, label style={text=red!70!black}]{1}{3}{xcomp}
\depedge[edge style={green!60!black, thick}, label style={text=green!60!black}]{9}{4}{mark}
\depedge[edge style={green!60!black, thick}, label style={text=green!60!black}]{7}{5}{det}
\depedge[edge style={red!70!black, thick}, label style={text=red!70!black}]{7}{6}{amod}
\depedge[edge style={red!70!black, thick}, label style={text=red!70!black}]{9}{7}{nsubj:pass}
\depedge[edge style={green!60!black, thick}, label style={text=green!60!black}]{9}{8}{aux:pass}
\depedge[edge style={red!70!black, thick}, label style={text=red!70!black}]{3}{9}{ccomp}
\depedge[edge style={green!60!black, thick}, label style={text=green!60!black}]{13}{10}{case}
\depedge[edge style={green!60!black, thick}, label style={text=green!60!black}]{13}{11}{det}
\depedge[edge style={red!70!black, thick}, label style={text=red!70!black}]{13}{12}{amod}
\depedge[edge style={red!70!black, thick}, label style={text=red!70!black}]{9}{13}{obl}
\depedge[edge style={green!60!black, thick}, label style={text=green!60!black}]{15}{14}{case}
\depedge[edge style={red!70!black, thick}, label style={text=red!70!black}]{13}{15}{nmod}
\end{dependency}

\vspace{0.1em}
\textbf{(a)} Universal Dependencies

\vspace{0.6em}

\begin{dependency}[theme=simple, edge style={thick}, label style={font=\tiny}]
\begin{deptext}[column sep=0.1cm, font=\small]
Let \& me \& say \& that \& the \& true \& revolutionary \& is \& guided \& by \& a \& great \& feeling \& of \& love \\
\end{deptext}
\deproot[edge unit distance=2ex]{1}{root}
\depedge[edge style={red!70!black, thick}, label style={text=red!70!black}]{1}{2}{obj}
\depedge[edge style={red!70!black, thick}, label style={text=red!70!black}]{1}{3}{comp:obj}
\depedge[edge style={red!70!black, thick}, label style={text=red!70!black}]{3}{4}{comp:obj}
\depedge[edge style={green!60!black, thick}, label style={text=green!60!black}]{7}{5}{det}
\depedge[edge style={red!70!black, thick}, label style={text=red!70!black}]{7}{6}{mod}
\depedge[edge style={red!70!black, thick}, label style={text=red!70!black}]{8}{7}{subj}
\depedge[edge style={green!60!black, thick}, label style={text=green!60!black}]{4}{8}{comp:obj}
\depedge[edge style={green!60!black, thick}, label style={text=green!60!black}]{8}{9}{comp:aux}
\depedge[edge style={red!70!black, thick}, label style={text=red!70!black}]{9}{10}{comp:obl}
\depedge[edge style={green!60!black, thick}, label style={text=green!60!black}]{13}{11}{det}
\depedge[edge style={red!70!black, thick}, label style={text=red!70!black}]{13}{12}{mod}
\depedge[edge style={green!60!black, thick}, label style={text=green!60!black}]{10}{13}{comp:obj}
\depedge[edge style={red!70!black, thick}, label style={text=red!70!black}]{13}{14}{mod}
\depedge[edge style={green!60!black, thick}, label style={text=green!60!black}]{14}{15}{comp:obj}
\end{dependency}

\vspace{0.2em}
\textbf{(b)} Surface-Syntactic UD

\caption{Dependency analysis of \emph{Let me say that the true revolutionary is guided by a great feeling of love} (Guevara, 1965). \textcolor{green!60!black}{Green arcs} = functional; \textcolor{red!70!black}{red arcs} = lexical. In~(a), functional elements depend on content words. In~(b), functional elements (auxiliaries, adpositions, complementizers) are heads.}
\label{fig:dep_example}
\end{figure*}

Figure~\ref{fig:dep_example} illustrates the functional--lexical distinction in both UD and SUD on a single sentence.

\subsection{UD to SUD Relation Mapping}

\begin{table}[!ht]
\centering
\resizebox{\columnwidth}{!}{
\begin{tabular}{llll}
\toprule
\textbf{UD rel.} & \textbf{SUD rel.} & \textbf{Class} & \textbf{Criterion} \\
\midrule
\texttt{det}  & \texttt{det}       & Func & same \\
\texttt{cc}   & \texttt{cc}        & Func & same \\
\texttt{clf}  & \texttt{clf}       & Func & same \\
\texttt{aux}  & \texttt{comp:aux}  & Func & label prefix \\
\texttt{cop}  & \texttt{comp:pred} & Func & label prefix \\
\texttt{case} & \texttt{comp:obj}  & Func & head = ADP \\
\texttt{mark} & \texttt{comp:obl}  & Func & head = SCONJ \\
\midrule
\texttt{nsubj}& \texttt{subj}      & Lex  & base label \\
\texttt{obj}  & \texttt{comp:obj}  & Lex  & head = VERB \\
\texttt{obl}  & \texttt{comp:obl}  & Lex  & head = VERB \\
\texttt{nmod} & \texttt{udep}      & Lex  & base label \\
\texttt{amod} & \texttt{mod}       & Lex  & base label \\
\texttt{advmod}& \texttt{mod}      & Lex  & base label \\
\texttt{conj}$^\dagger$ & \texttt{conj:*}    & Lex  & base label \\
\bottomrule
\end{tabular}
}
\caption{UD $\to$ SUD relation mapping and functional/lexical classification. SUD \texttt{comp:obj/obl} is disambiguated via the head's UPOS tag. $^\dagger$Excluded under Scenarios~B--D in the sensitivity analysis.}
\label{tab:sud_mapping}
\end{table}

\subsection{Detailed Sensitivity Analysis}

Our main analysis classifies 7 base relation types as functional and 23 as lexical. Several lexical relations are arguably not prototypical syntactic dependencies: \texttt{conj} (mean MDD~5.45) is inflated by UD's chain analysis; \texttt{parataxis} (6.63) and \texttt{discourse} (3.12) are discourse-level; \texttt{flat} (1.41), \texttt{fixed} (1.15), and \texttt{compound} (1.23) are MWE-internal.

We tested three alternative groupings:

\begin{itemize}
\item \textbf{Scenario~B ($-$conj):} Excluding \texttt{conj}. Lexical MDD drops to 2.66 ($\pm$0.56); func $<$ lex in 122/122 languages ($d = 1.89$).
\item \textbf{Scenario~C (core syntax):} Restricting lexical to core arguments and modifiers (\texttt{nsubj}, \texttt{obj}, \texttt{iobj}, \texttt{obl}, \texttt{nmod}, \texttt{amod}, \texttt{advmod}, \texttt{advcl}, \texttt{acl}, \texttt{xcomp}, \texttt{ccomp}, \texttt{csubj}, \texttt{nummod}). Lexical MDD = 2.65 ($\pm$0.66); func $<$ lex in 121/122 ($d = 1.65$).
\item \textbf{Scenario~D (strictest):} Removing \texttt{mark} and \texttt{cc} from functional; \texttt{advmod} and \texttt{compound} from lexical. Func MDD = 1.40 ($\pm$0.25), lex MDD = 2.75 ($\pm$0.69); gap \emph{widens} ($d = 2.08$).
\end{itemize}

\textbf{Pronoun dependents.} Pronoun-headed lexical arcs are shorter (MDD = 2.34 $\pm$ 0.97) than non-pronoun (2.93 $\pm$ 0.72), in 81\% of treebanks. However, pronouns are only 12\% of lexical tokens; excluding them \emph{raises} lexical MDD, widening the gap.

\textbf{Nucleus-based distance.} Computing distances counting only content-word positions reduces lexical MDD from 2.87 to 2.12, but the asymmetry is preserved (2.12 still exceeds functional MDD 1.71; $r = 0.78$ with standard measure).

\textbf{Sample-size stability.} Bootstrap resampling on 8 diverse languages shows the gap is detectable at 100 sentences (8/8 languages) and even at 50 sentences (7/8). All languages with $\geq$200 sentences individually confirm lex~$>$~func.

\section*{Ethics Statement and AI Disclosure}\label{sec:ethics}

We fully disclose that this paper was produced using an AI system. The research idea, all analysis code, and the entire text were generated by Claude Opus 4.6 (Anthropic), operating in agentic mode via GitHub Copilot Chat in VS~Code. The human author provided prompts and accepted or rejected proposed work at each step, but did not originate the central research question, write any code, or draft any prose beyond the prompts themselves. The author's contribution was verification and editorial oversight rather than generation of the core ideas, code, or prose.

The project was inspired by an experiment by D.~Yanagizawa-Drott, who prompted an LLM to write a macroeconomics job market paper and publicly reflected on the implications (\url{https://x.com/YanagizawaD/status/2022034189395407093}). This paper began with an analogous prompt: ``\emph{I always dreamt of becoming a syntactician and a typologist one day. I want to submit a paper to the UDW workshop. Can you help me?}'' The functional--lexical distinction that became the core contribution emerged during the AI's iterative exploration of the data, not from the human author. Prior work was identified using Google Scholar Labs and provided to the system for integration.

The reviews by anonymous UDW~2026 reviewers and by Sylvain Kahane were essential for improving the final version; reviewer feedback was given to the AI, which implemented the revisions. The code, analyses, and results were checked by the author, and the final paper benefited substantially from this review process.

We believe the scientific community deserves full transparency about how research is produced. As Yanagizawa-Drott noted, in a world of machine-speed generated papers we face a dilemma: either this is real research that requires expert verification, or it is not, and we have polluted the information environment. We hope this disclosure contributes to an honest conversation about the role of AI in scientific work. This work was produced with substantial AI assistance, and readers should nonetheless be aware of this production process when evaluating the claims.

An open question is whether AI-generated hypotheses and text, curated and verified by a human, should count as a scientific result, and whether this may become a standard category of research output. In this case, both the author and the reviewers judged the findings scientifically interesting enough to merit revision and discussion. More broadly, this process raises an old question about novelty: can LLMs create genuinely new ideas, can humans, or are both primarily recombining from a finite space of possibilities, as in Borges' ``Library of Babel''? We do not claim to resolve this question here, but we consider it central for future norms of authorship, credit, and evaluation.

\section{Bibliographical References}\label{sec:reference}

\bibliographystyle{lrec2026-natbib}
\bibliography{references}

\end{document}